\documentclass{article}
\usepackage{ais25}
\usepackage[T1]{fontenc}
\usepackage{blindtext}
\usepackage{xcolor}
\usepackage{circuitikz}
\usepackage{tikz}
\usepackage{hyperref}
\usepackage{url}
\usepackage{array}
\usepackage{tabu}
\usepackage[]{caption}
\usepackage{graphicx}

\defcitealias{dmi2023klimaatlas}{[DMI, 2023]}

\title{%
Incorporating Quality of Life in Climate Adaptation Planning via Reinforcement Learning
}
\author{%
\name{Miguel Costa} \addr{Technical University of Denmark} \email{migcos@dtu.dk} \\
\name{Arthur Vandervoort} \addr{Technical University of Denmark} \email{apiva@dtu.dk} \\
\name{Martin Drews} \addr{Technical University of Denmark}  \email{mard@dtu.dk} \\ 
\name{Karyn Morrissey}  \addr{Univ. of Galway} \email{karyn.morrissey@universityofgalway.ie} \\
\name{Francisco C. Pereira} \addr{Technical University of Denmark}  \email{camara@dtu.dk} }

\begin{document}
\maketitle

\begin{abstract}
Urban flooding is expected to increase in frequency and severity as a consequence of climate change, causing wide-ranging impacts that include a decrease in urban Quality of Life (QoL). 
Meanwhile, policymakers must devise adaptation strategies that can cope with the uncertain nature of climate change and the complex and dynamic nature of urban flooding. Reinforcement Learning (RL) holds significant promise in tackling such complex, dynamic, and uncertain problems.
Because of this, we use RL to identify which climate adaptation pathways lead to a higher QoL in the long term. We do this using an Integrated Assessment Model (IAM) which combines a rainfall projection model, a flood model, a transport accessibility model, and a quality of life index.
Our preliminary results suggest that this approach can be used to learn optimal adaptation measures and it outperforms other realistic and real-world planning strategies. 
Our framework is publicly available: \url{https://github.com/MLSM-at-DTU/maat_qol_framework}.

\end{abstract}

\begin{keywords}
Quality of Life, Climate Adaptation, Pluvial Flooding, Reinforcement Learning, Transportation
\end{keywords}

\section{Introduction}
\label{sec:introduction}

The drive to measure Quality of Life (QoL) in cities has given urban planners and policymakers the tools to identify needs not typically captured by traditional economic metrics \citep{maransQualityLifeLargescale2024a, atkinson2013beyond, lee2016transportation}. We define QoL as the set of amenities accessible to a given person at a given time, measuring the ``circumstances of a person's life rather than his or her response to those circumstances`` \citep{diener2006guidelines}. As urban flooding looks to become more prevalent in the coming century, adapting cities to flooding while maintaining QoL should be a critical priority for urban planners \citep{ipcc2023climate}.

Climate change will significantly impact mobility and access to locations and services. High-impact weather events, including extreme rainfall \citep{ipcc2023climate}, are projected to rise in frequency and intensity, increasing disruptions caused by urban flooding \citep{olesen2014fremtidige}. Flooding disrupts economic and social activities \citep{hammond2015urban}, restricting individuals’ access and, consequently, affecting QoL. These disruptions have both immediate and long-term implications for urban resilience and accessibility.

Copenhagen is among Denmark’s most vulnerable cities to flooding \citep{prallComprehensiveApproachAssessing2024}. To make the city resilient while maintaining or improving QoL, policymakers must choose effective adaptation strategies to implement over time. Given the complex and dynamic nature of climate change and urban flooding, reinforcement learning (RL) presents a promising approach to optimise policy decisions \citep{gilbert2022choicesrisksrewardreports}. RL can be used to discover which adaptation strategies balance long- and short-term quality of life with economic costs.

Thus, we aim to answer the following question: ``\textit{What is the sequence of climate adaptation measures that lead to a higher quality of life in the long term?}'' Seeking to minimise the impacts of climate-related pluvial flooding on quality of life, we explore how different climate adaptation strategies impact accessibility to key services and use RL as a tool to identify adaptation sequences that maintain current levels of QoL under uncertain climate projections. Though we are focusing on Copenhagen's inner city as a case study, our framework can easily be expanded to other cities, climate-related impacts (e.g., coastal flooding or wildfires), and contexts (e.g., subjective wellbeing or mobility justice).

\section{Methodology}
\label{sec:introduction}
\begin{figure}[htb]
    \vspace{-12pt}
    \resizebox{0.95\linewidth}{!}{%
        \tikzset{every picture/.style={line width=0.75pt}} 
        \begin{tikzpicture}[x=0.65pt,y=0.65pt,yscale=-.9,xscale=1]
                


\draw [line width=2.25] (25,25) -- (705,25) -- (705,150) -- (25,150) -- cycle  ;

\draw (705,96) -- (715,96) -- (715,200) -- (440,200) ;
\draw [shift={(440,200)}, rotate = 360] [fill={rgb, 255:red, 0; green, 0; blue, 0 }  ][line width=0.08]  [draw opacity=0] (8,-4) -- (0,0) -- (8,4) -- cycle ;
\draw (705,83) -- (724,84) -- (724,210) -- (440,210) ;
\draw [shift={(440,210)}, rotate = 360] [fill={rgb, 255:red, 0; green, 0; blue, 0 }  ][line width=0.08]  [draw opacity=0] (8,-4) -- (0,0) -- (8,4) -- cycle ;
\draw (255,206) -- (13,206) -- (13,90) -- (20,90) ;
\draw [shift={(25,90)}, rotate = 180] [fill={rgb, 255:red, 0; green, 0; blue, 0 }  ][line width=0.08]  [draw opacity=0] (8,-4) -- (0,0) -- (8,4) -- cycle ;

\draw (530,219) node [font=\small] [align=left] {\textbf{State}};
\draw (530,189) node [font=\small] [align=left] {\textbf{Reward}};
\draw (136,197) node [font=\small] [align=left] {\begin{minipage}[lt]{100.pt}\setlength\topsep{0pt}
\begin{center}
{\footnotesize (Adaptation measures)}
\end{center}

\end{minipage}};
\draw (650,11) node [align=left] {Environment};
\draw (136.25,178.9) node [font=\small] [align=left] {\begin{minipage}[lt]{32.83pt}\setlength\topsep{0pt}
\begin{center}
\textbf{Actions}
\end{center}
\end{minipage}};

\draw    (40,50) -- (200,50) -- (200,130) -- (40,130) -- cycle  ;
\draw (120,90) node   [align=left] {\begin{minipage}[lt]{99.08pt}\setlength\topsep{0pt}
\begin{center}
\textbf{Rainfall Projection Model}\\{\scriptsize{\citep{dmi2023klimaatlas}}}
\end{center}
\end{minipage}};

\draw    (225,50) -- (335,50) -- (335,130) -- (225,130) -- cycle  ;
\draw (280,90) node   [align=left] {\begin{minipage}[lt]{99.08pt}\setlength\topsep{0pt}
\begin{center}
\textbf{Flood Model}\\
{\scriptsize{\citep{scalgo}}}
\end{center}
\end{minipage}};

\draw    (360,50) -- (520,50) -- (520,130) -- (360,130) -- cycle  ;
\draw (440,90) node   [align=left] {\begin{minipage}[lt]{99.08pt}\setlength\topsep{0pt}
\begin{center}
\textbf{Transport Accessibility}\\
{\scriptsize{(Cumulative-based accessibility)}}
\end{center}
\end{minipage}};

\draw    (545,50) -- (690,50) -- (690,130) -- (545,130) -- cycle  ;
\draw (615,90) node   [align=left] {\begin{minipage}[lt]{99.08pt}\setlength\topsep{0pt}
\begin{center}
\textbf{Quality of Life Index}\\
{\scriptsize{(Transport-based\\Quality of Life)}}
\end{center}
\end{minipage}};

\draw  [draw opacity=0]  (225.35,178.7) -- (407.35,178.7) -- (407.35,232.7) -- (225.35,232.7) -- cycle  ;
\draw (350,205) node   [align=left] {\begin{minipage}[lt]{120.84pt}\setlength\topsep{0pt}
\begin{center}
Reinforcement Learning Agent
\end{center}
\end{minipage}};

\draw    (200,90) -- (225,90) ;
\draw [shift={(225,90)}, rotate = 180] [fill={rgb, 255:red, 0; green, 0; blue, 0 }  ][line width=0.08]  [draw opacity=0] (8.93,-4.29) -- (0,0) -- (8.93,4.29) -- cycle ;
\draw    (335,90) -- (360,90) ;
\draw [shift={(360,90)}, rotate = 180] [fill={rgb, 255:red, 0; green, 0; blue, 0 }  ][line width=0.08]  [draw opacity=0] (8.93,-4.29) -- (0,0) -- (8.93,4.29) -- cycle ;
\draw    (520,90) -- (545,90) ;
\draw [shift={(545,90)}, rotate = 180] [fill={rgb, 255:red, 0; green, 0; blue, 0 }  ][line width=0.08]  [draw opacity=0] (8.93,-4.29) -- (0,0) -- (8.93,4.29) -- cycle ;

        \end{tikzpicture}
    }
    \vspace{-12pt}
    \caption{IAM using RL to learn what the best sequence of adaptation policies are to maximise quality of life under climate changing scenarios.}
    \label{fig:iam_framework}
\end{figure}
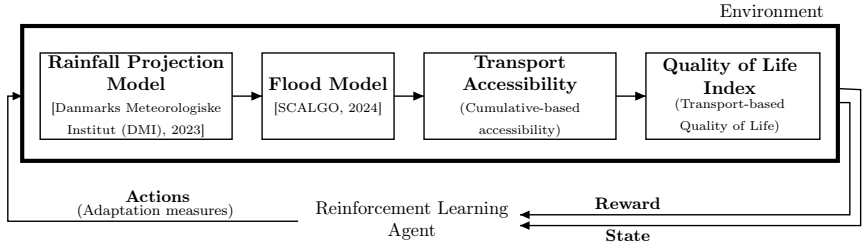

We frame our approach as an Integrated Assessment Model (IAM) that includes a rainfall model, 2) a flood model, 3) a transport accessibility component, and 4) a quality of life index model. 
Figure \ref{fig:iam_framework} provides an overview of our framework.

\vspace{-12pt}

\subsection{Rainfall Projection and Flood Model}

To simulate rainfall events, we sampled precipitation intensities based on probability distributions built from the Danish Meteorological Institute's Climate Atlas \citetalias{dmi2023klimaatlas} for 2023--2100. For simplicity, we assumed the projected rainfall intensity to be equal to the accumulated daily rainfall. After sampling a rainfall event (i.e., amount of rainfall), we modelled the associated urban flood. Urban pluvial flooding is often caused by intense short duration precipitation (cloudbursts, from minutes to a few hours). For this, we used SCALGO Live \citep{scalgo}, a event-based tool for watershed delineation, flood depth, and flow direction modelling based on digital terrain data. For any rainfall event, water was distributed according to its intensity and terrain properties. In the end, water depths of resulting floods were mapped in Copenhagen.



\subsection{Transport Accessibility and Quality of Life Index}

We define QoL as the weighted sum of the per-capita number of amenities accessible within a reasonable travel time from residential locations \citep{dobrowolskaMappingUrbanWellbeing2024}. Taking this approach enables us to model QoL at a granular level, which allows us to identify the effects of flooding across (and within) urban administrative boundaries where QoL indicators are typically gathered \citep{maransQualityLifeLargescale2024a}.

Our QoL index is adapted from previous work on QoL by \citet{dobrowolskaMappingUrbanWellbeing2024}. First, we compute the per-capita number of accessible points of interest (POI) of a given category for a location. We then take the weighted sum of all POIs for a given location, where weights are determined using a logistic regression model that predicts satisfaction with life in cities from \citep{ec/dgregioQualityLifeEuropean2023}. This yields an index that maps the relative density of amenities per capita for a given area, weighted by the relative importance of those amenities based on survey data.

\subsection{Reinforcement Learning}

We posit to learn the best sequence of adaptation policies that maximise QoL using RL. RL uses an agent-based approach to interact with the above environment by taking an action (adaptation measure) and maximising a (delayed) reward function \citep{sutton2018reinforcement}, learning to balance trade-offs between competing actions and input uncertainty in the environment. 

In this work, we propose eight actions that can be implemented in each area of Copenhagen. These include increasing road drainage and permeable paving solutions. Actions change the environment, directly changing transport and indirectly affecting QoL. We define the reward function to optimize for as:\vspace{-6pt}
\begin{equation}
    R = \sum_{i} 
    \beta_{Q} Q_{i} + 
    \beta_{A} A_{i} +
    \beta_{M} M_{i}
\end{equation}
where $Q_{i}$ corresponds to Quality of Life index at the $i$-th zone, $A_{i}$ is the cost of applying an action and $M_{i}$ its maintenance cost over time. This reward function is highly customisable. Users can choose different $\beta$ weights depending on competing priorities and trade-offs between QoL and economic costs. 

To learn which action to perform at a given time, our RL agent takes an action and collects information on the state of our digital city. Over time, the agent learns the best sequence of policies that maximise the cumulative reward.

\vspace{-12pt}

\section{Preliminary Results and Discussion}
\label{sec:results}
We setup our IAM using Python, Gymnasium interface \citep{towers_gymnasium_2023}, Stable-Baseline3 \citep{stable_baselines3}, and PPO \citep{huang2020closer, schulman2017proximal}. As a preliminary case study, we conduct an experiment in Copenhagen's inner city (consisting of 29 zones) by setting the time horizon between 2023--2100, and setting $\beta$ weights that prioritise QoL over economic costs. We now present preliminary results for five distinct seed runs to allow for different weather projections and robustness.

Figure~\ref{fig:results} showcases the results comparing our learnt (Optimal Policy) with five feasible baselines (see Appendix \ref{sec:appendix}). Our results show that the RL framework achieves better rewards compared to all baselines. The similarity between the random and the learnt policy can be explained by our experimental setup, which places heavier emphasis on QoL. This reduces the relative importance of action costs, leading to a similar cumulative reward to the random policy. Crucially, the sequence of actions taken by the agent is different to the random policy, leading to small increases in cumulative QoL and small decreases in cumulative action costs. Future experiments with different weightings are likely to yield further improvements.

\begin{figure}[htb]
\centering
    \begin{minipage}{.45\linewidth}
        \centering
        \includegraphics[width=0.98\textwidth]{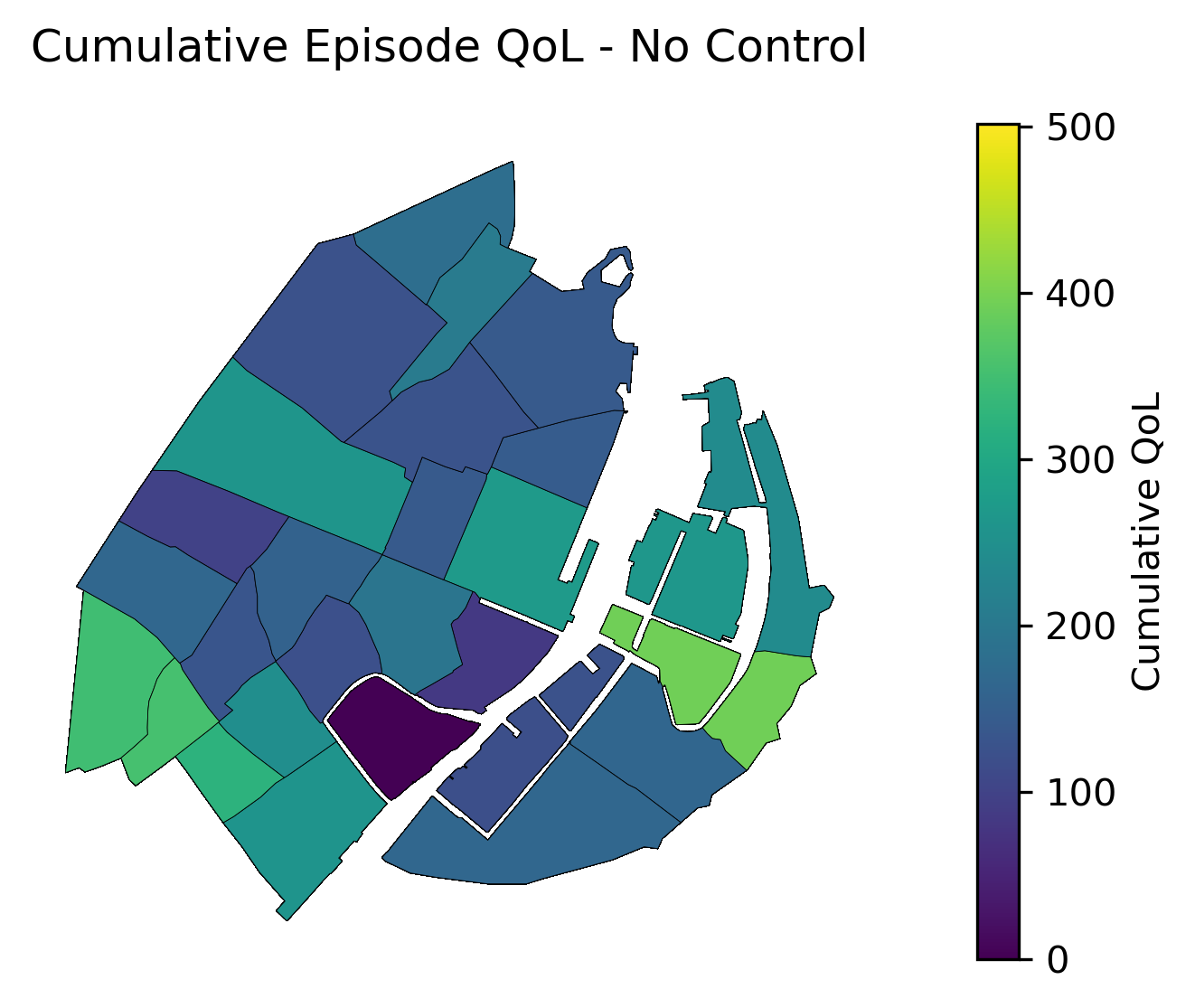}
    \end{minipage}
    \begin{minipage}{.45\textwidth}
        \centering
        \includegraphics[width=0.98\textwidth]{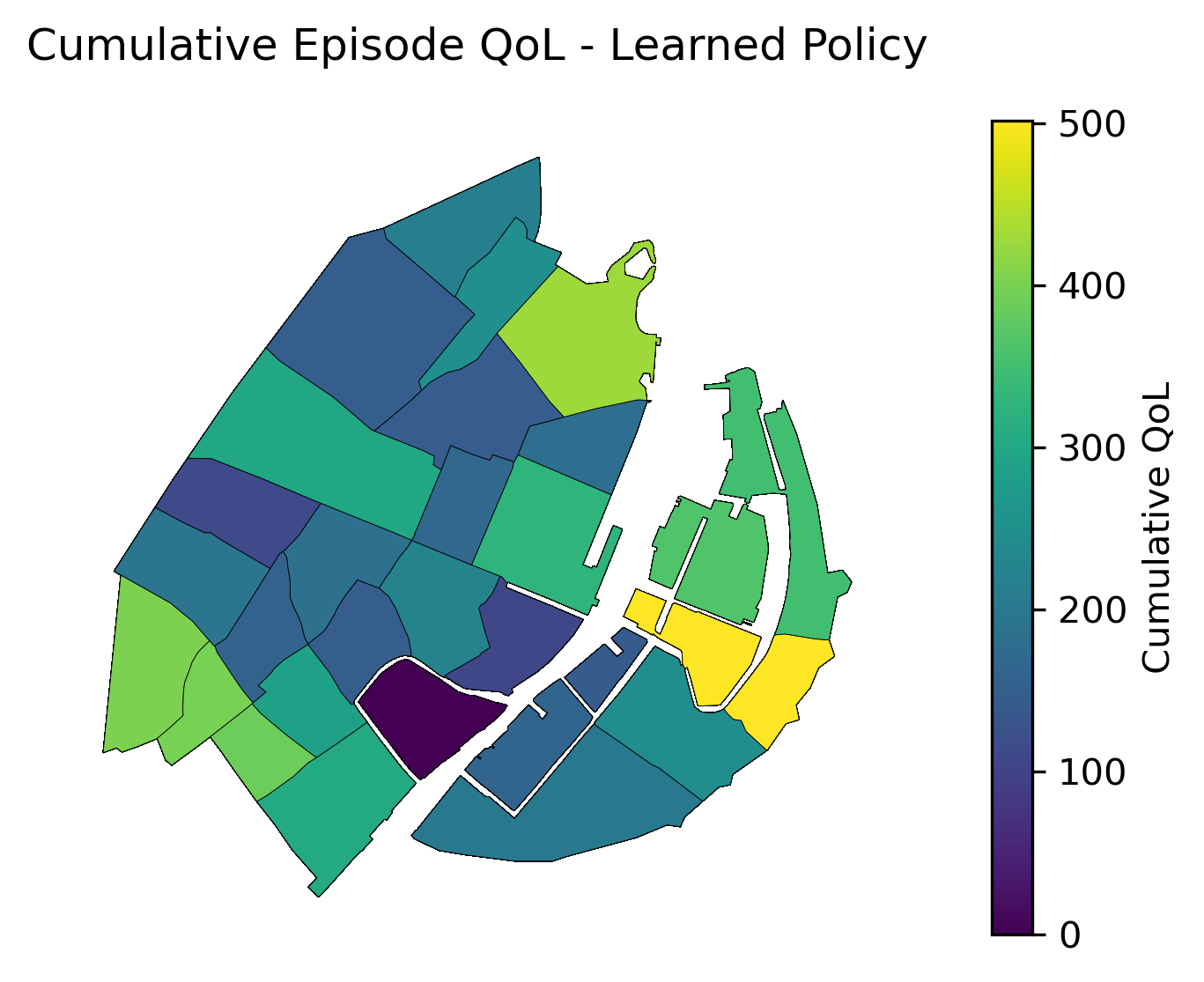}
    \end{minipage} \\
    \begin{minipage}{.45\textwidth}
        \centering
        \includegraphics[width=0.98\textwidth]{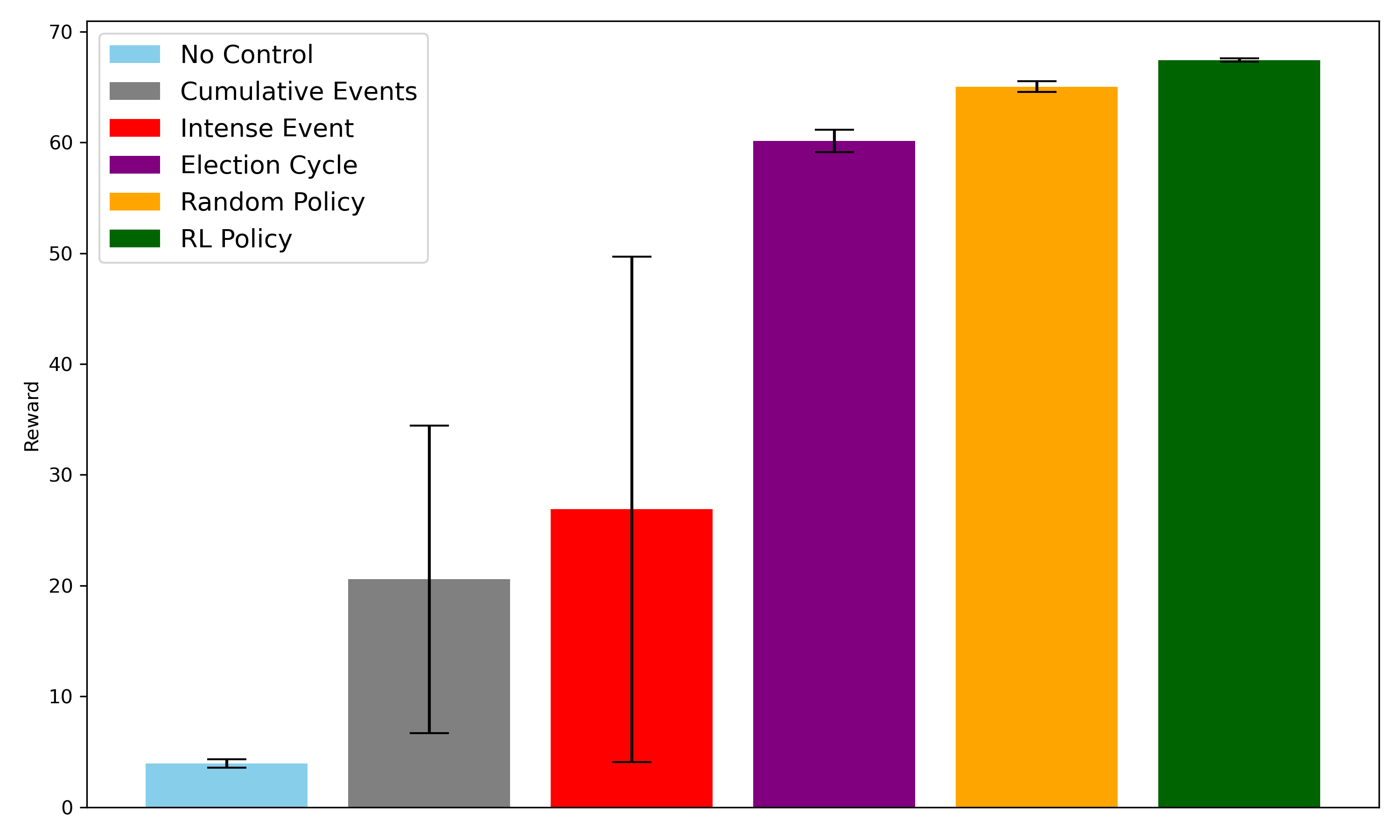}
    \end{minipage}
    \begin{minipage}{.45\textwidth}
        \centering
        \includegraphics[width=0.98\textwidth]{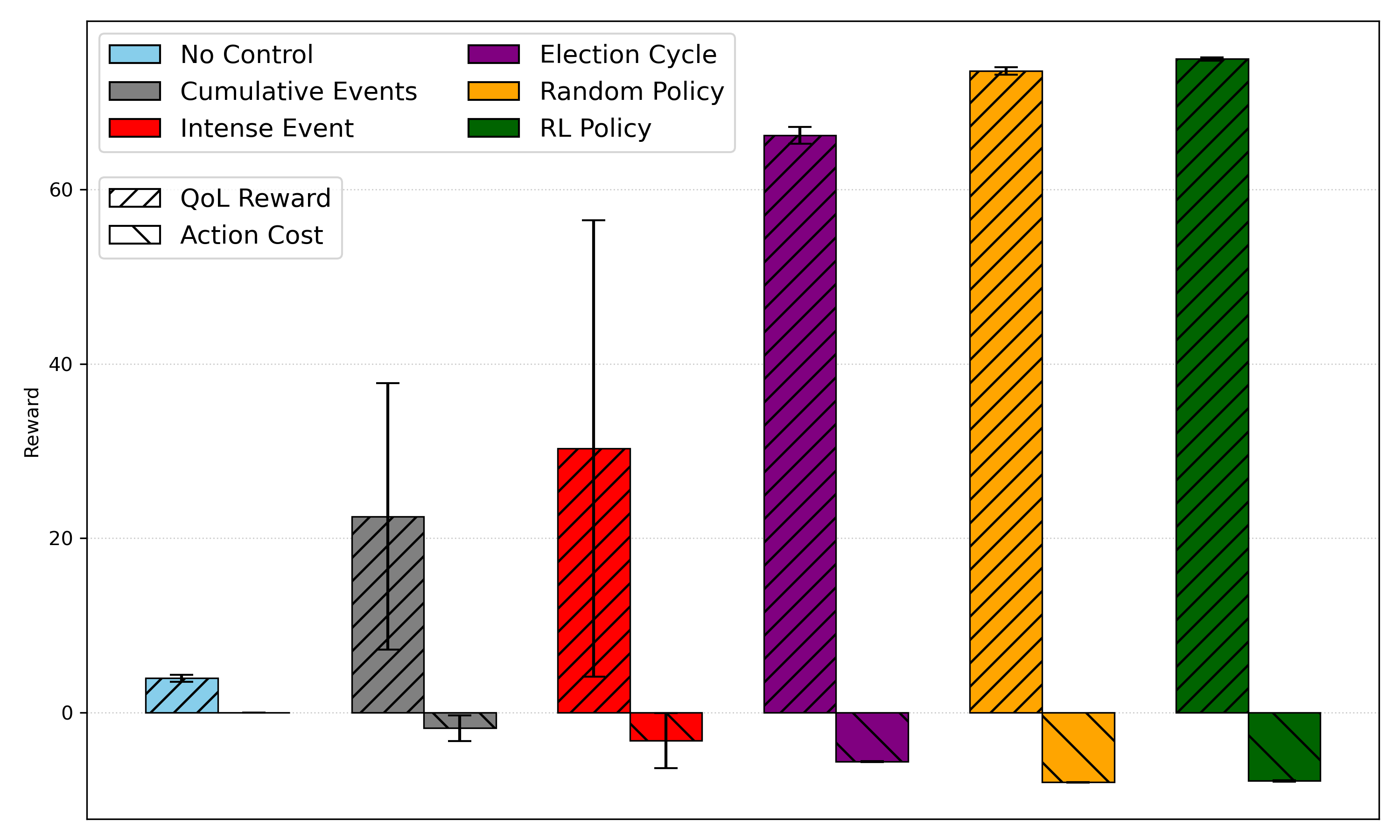}
    \end{minipage}
\caption{Top row: Side-by-side comparison of No Control and Learnt Policy result by city zone. Bottom row: Comparison of total reward between Learnt Policy and baselines (left), and between reward components (right).}
\label{fig:results}
\end{figure}

Our results demonstrate the feasibility of using RL to construct IAMs and identify policy pathways focused on minimising QoL loss, in contrast to the cost-minimisation or welfare-optimisation approaches typical used in IAMs \citep{weyant_contributions_2017}. Though these results are promising, IAMs (with or without RL) feature multiple quantifiable and unquantifiable uncertainties that are important to consider \citep{beckEpistemicEthicalPolitical2016}. Moving forward, we plan to continue to investigate the effects of normative choices in models such as these, including testing the sensitivity of the model to different $\beta$ weights for QoL (representing the 'value' of QoL relative to economic costs) and using sufficientarian metrics (e.g. Foster-Greer-Thorbecke indices \citep{karnerAdvancesPitfallsMeasuring2024a}). Besides this, we also intend to extend our framework to the full city of Copenhagen.

\section*{Acknowledgements}
This work was supported by a research grant (VIL57387) from VILLUM FONDEN.

\bibliography{ais}

\appendix
\section{Appendix A}
\label{sec:appendix}

We compare the learnt (Optimal Policy) reward versus five other baselines: 
\begin{itemize}
    \item No Control: no actions are implemented
    \item Cumulative Response: adaptation starts when 3 10-year return period rainfall events occur within a 10-year period
    \item Intense Event Response: adaptation starts after a 20-year return period event occurs
    \item Election Cycle: adaptation occurs every 4 years
    \item Random Control: actions are taken randomly
\end{itemize}

\end{document}